\documentclass[5p,times]{elsarticle}

\usepackage{graphicx}
\usepackage{amsmath}
\usepackage{booktabs}
\usepackage{array}
\usepackage{tabularx}
\usepackage{microtype}
\usepackage{caption}
\usepackage{url}
\usepackage[hidelinks]{hyperref}
\begin{document}
\begin{frontmatter}
\title{DPTM-DT: Dual-Pretrained Transformer Multitask Representation Learning for Drug-Target Prediction}
\author[beihang]{Ge Kong}
\address[beihang]{School of Biological Science and Medical Engineering, Beihang University, Beijing 100191, China\\
E-mail: \textnormal{\href{mailto:gekong@buaa.edu.cn}{gekong@buaa.edu.cn}}}
\begin{abstract}
Drug-target relation prediction supports candidate screening, drug repositioning, and mechanism analysis. Existing models often use incomplete drug or protein representations, model cross-modal interactions shallowly, or train affinity regression and interaction classification separately, although these tasks describe closely related views of the same drug-target pair. This paper presents DPTM-DT, a dual-pretrained Transformer framework for multitask drug-target prediction. DPTM-DT combines GROVER molecular graph embeddings, ESM protein language-model embeddings, and CTD physicochemical descriptors, then exchanges drug-target information through bidirectional cross-modal attention. A shared pair representation is used for continuous affinity regression, high-affinity binary classification, and six-level affinity classification. Experiments on Davis and KIBA cover random 80/20 and DeepDTA-style standard splits. On the random 80/20 split, DPTM-DT achieves MSE/CI values of 0.193/0.917 on Davis and 0.120/0.918 on KIBA. It also reports binary AUPR/MCC values of 0.727/0.654 and 0.798/0.689, and six-class Macro-F1/Top-2 values of 0.800/0.932 and 0.815/0.962 on Davis and KIBA, respectively. Across the reported regression, binary classification, and multiclass classification settings, DPTM-DT achieves the best overall performance among the compared methods. Results under the standard split show the same relative trend. Ablations indicate that dual target representation, gated fusion, and cross-modal attention each contribute to the final performance. Code and supplementary materials are available at: \url{anonymous.4open.science/r/DPCM-DT-74E0}.
\end{abstract}
\begin{keyword}
drug-target prediction \sep binding affinity prediction \sep cross-modal learning \sep multitask learning \sep pretrained representation
\end{keyword}
\end{frontmatter}

\section{Introduction}

Drug-target interaction (DTI) and drug-target affinity (DTA) prediction are central tasks in computational drug discovery. Experimental validation is reliable but expensive, while the number of possible compound-protein pairs is large. Computational models can therefore help prioritize candidates before wet-lab testing. Recent deep methods learn from SMILES strings, molecular graphs, and protein sequences, improving on earlier descriptor- and similarity-based approaches \cite{bagherian2021,gnndtasurvey,deepdta,widedta,glcn,tefdta}.

Existing DTI and DTA models still face three limitations. First, a single representation rarely captures both molecular topology and protein sequence or physicochemical information. Second, simple concatenation cannot fully describe the correspondence between drug and target modalities. Third, regression, binary classification, and affinity-level classification are related views of the same pairwise relation but are often optimized separately. DPTM-DT addresses these limitations through dual pretrained representations, cross-modal Transformer interaction, and multitask prediction.

DPTM-DT addresses these issues with a multitask architecture in which representation learning, cross-modal interaction, and task supervision are trained around the same drug-target pair. The design has three parts: dual pretrained drug and protein representations, bidirectional drug-target information exchange, and a shared pair representation for multiple affinity-oriented objectives. The paper makes three contributions: (1) a dual-pretrained representation module that combines GROVER, ESM, and CTD features; (2) a cross-modal joint interaction module for pairwise drug-target matching; and (3) an evaluation protocol covering affinity regression, binary interaction classification, and six-class affinity-level classification on Davis and KIBA under two split protocols.

\section{Related Work}

Drug-target prediction has moved from handcrafted descriptors and similarity kernels toward end-to-end representation learning. DeepDTA encodes SMILES strings and protein sequences with convolutional networks, showing that affinity prediction can be learned directly from raw symbolic inputs \cite{deepdta}. WideDTA extends this idea with word-level drug and protein representations, including motifs, domains, and molecular substructures \cite{widedta}. Graph-based models such as GLCN-DTA further exploit molecular topology, while TEFDTA combines Transformer encoding with molecular fingerprints for affinity prediction \cite{glcn,tefdta}. Complementary feature-fusion DTA models have also explored multi-level sequence features, hybrid fingerprints with protein N-grams, and E3FP-based multimodal molecular representations \cite{mlffdta,trifpngram,e3mmdta}. Recent methods continue this trend through multimodal fusion, dual attention, cross-scale graph contrastive learning, and multi-task co-attention \cite{kanmodti,dacmf,crossscale,multitaskcoattn,msidti,ammvf,kgmacnf,mutualattn}. Reviews of graph neural networks and Transformer models in drug discovery also suggest that representation quality and evaluation protocol design remain central issues for DTI and DTA models \cite{gnndtasurvey,transformerdrugreview}. DPTM-DT follows these representation-learning models while emphasizing protein-side complementarity and explicit drug-target interaction.

Recent pretrained models provide a natural way to strengthen each modality before drug-target matching. GROVER learns molecular graph semantics from large-scale unlabeled molecules, and ESM learns contextual protein sequence representations from evolutionary-scale protein data \cite{grover,esm2}. In parallel, CTD descriptors remain useful because they summarize physicochemical protein properties in a compact and interpretable form \cite{ctd}. More broadly, foundation models and molecular representation learning have become increasingly important for drug discovery because they can transfer reusable structural and sequence information across downstream tasks \cite{foundationmodels,molrepr}. Recent studies have also applied graph attention and pretrained language models to biomedical omics, protein localization, and RNA localization prediction tasks \cite{gatomics,slpt5,mrnaloc}. These studies indicate that the field is moving toward richer molecular and protein representations, but many evaluations still emphasize a single prediction target or a single split protocol. DPTM-DT follows this direction by combining pretrained molecular and protein embeddings with CTD features, then using cross-modal attention and multitask heads to evaluate the same pair representation under regression, binary classification, and affinity-level classification.

\section{Method}

\subsection{Overall Pipeline}

\begin{figure*}[t]
\centering
\includegraphics[width=0.96\textwidth]{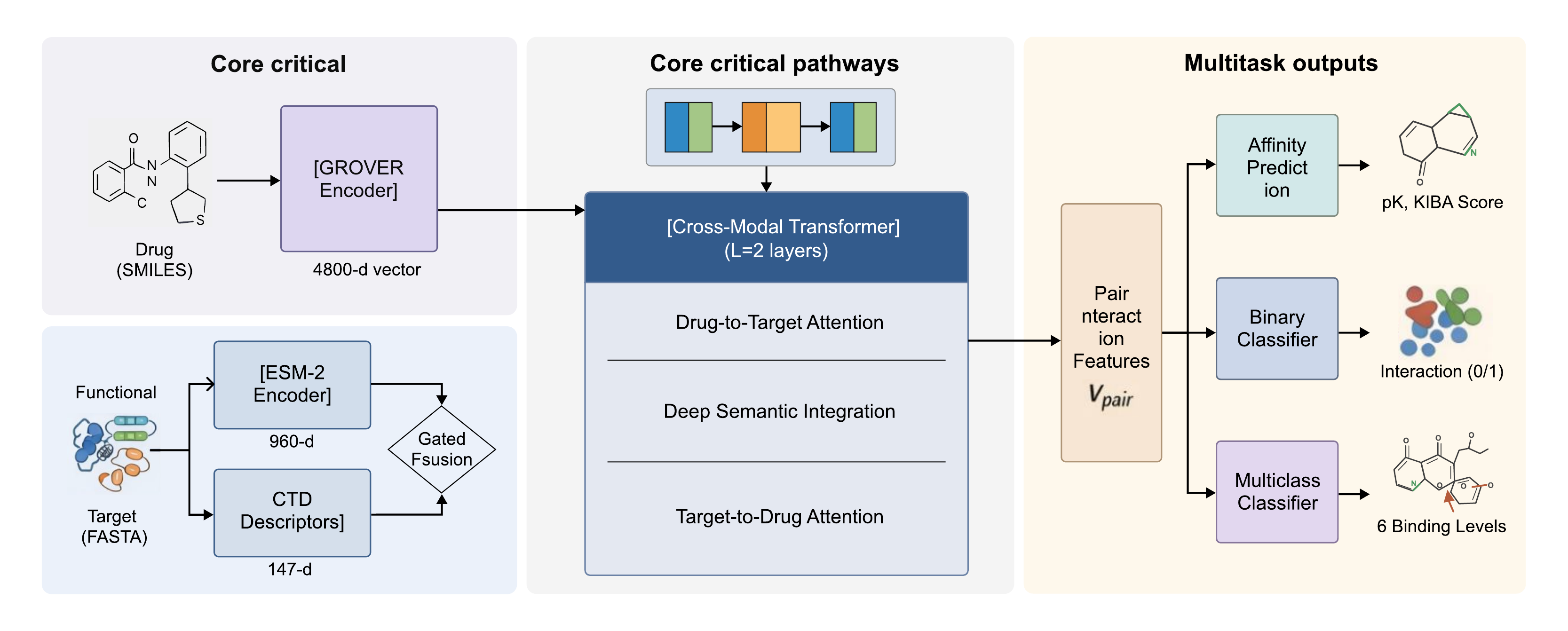}
\caption{Overall architecture of DPTM-DT. The model encodes drug SMILES with GROVER, encodes target FASTA sequences with ESM-2 and CTD descriptors, fuses target-side features, performs bidirectional cross-modal Transformer interaction, and predicts affinity regression, binary interaction, and six-level affinity classes from the shared pair representation.}
\label{fig:framework}
\end{figure*}

DPTM-DT models a sample as a drug-target pair $(d_i,p_i)$ with a continuous affinity label $a_i$. As shown in Fig.~\ref{fig:framework}, the model first extracts drug-side and target-side representations, maps all modalities into a shared 256-dimensional space, performs target-side feature fusion, updates the pair through cross-modal interaction, and finally predicts three task outputs. The three supervised tasks reuse the same pair representation: continuous affinity regression, binary interaction classification, and six-class affinity-level prediction.

\subsection{Drug and Target Representation}

On the drug side, DPTM-DT uses GROVER to encode molecular graph information. The resulting drug vector is denoted as $f_i^{drug}\in\mathbb{R}^{4800}$. This representation captures molecular topology, local atom-bond patterns, and graph-level chemical semantics learned from large-scale molecular pretraining.

On the target side, the model uses two complementary sources. ESM provides a protein language-model embedding $f_i^{esm}\in\mathbb{R}^{960}$, encoding sequence context and long-range dependencies. CTD provides a 147-dimensional descriptor $f_i^{ctd}\in\mathbb{R}^{147}$ based on composition, transition, and distribution statistics over physicochemical properties \cite{ctd}. These properties include hydrophobicity, polarity, charge, van der Waals volume, polarizability, solvent accessibility, and secondary-structure tendency.

All input modalities are projected into a common hidden space:
\begin{equation}
v_i^m=\Pi_m(f_i^m),\quad m\in\{drug,esm,ctd\},
\end{equation}
where $\Pi_m$ is a modality-specific projection with nonlinearity, dropout, and layer normalization. The target branch then fuses ESM and CTD. The full model uses gated fusion:
\begin{equation}
g_i=\sigma(W_g[v_i^{esm};v_i^{ctd}]+b_g),
\end{equation}
\begin{equation}
v_i^{target}=\mathrm{LN}(g_i\odot v_i^{esm}+(1-g_i)\odot v_i^{ctd}).
\end{equation}
Ablation variants compare this design with ESM-only, CTD-only, and direct concatenation. The gated design follows the same general motivation as gated multimodal fusion, where the network learns how much each modality contributes rather than using a fixed combination rule \cite{gated}.

\subsection{Cross-Modal Interaction and Pair Features}

\begin{figure}[t]
\centering
\includegraphics[width=0.96\columnwidth]{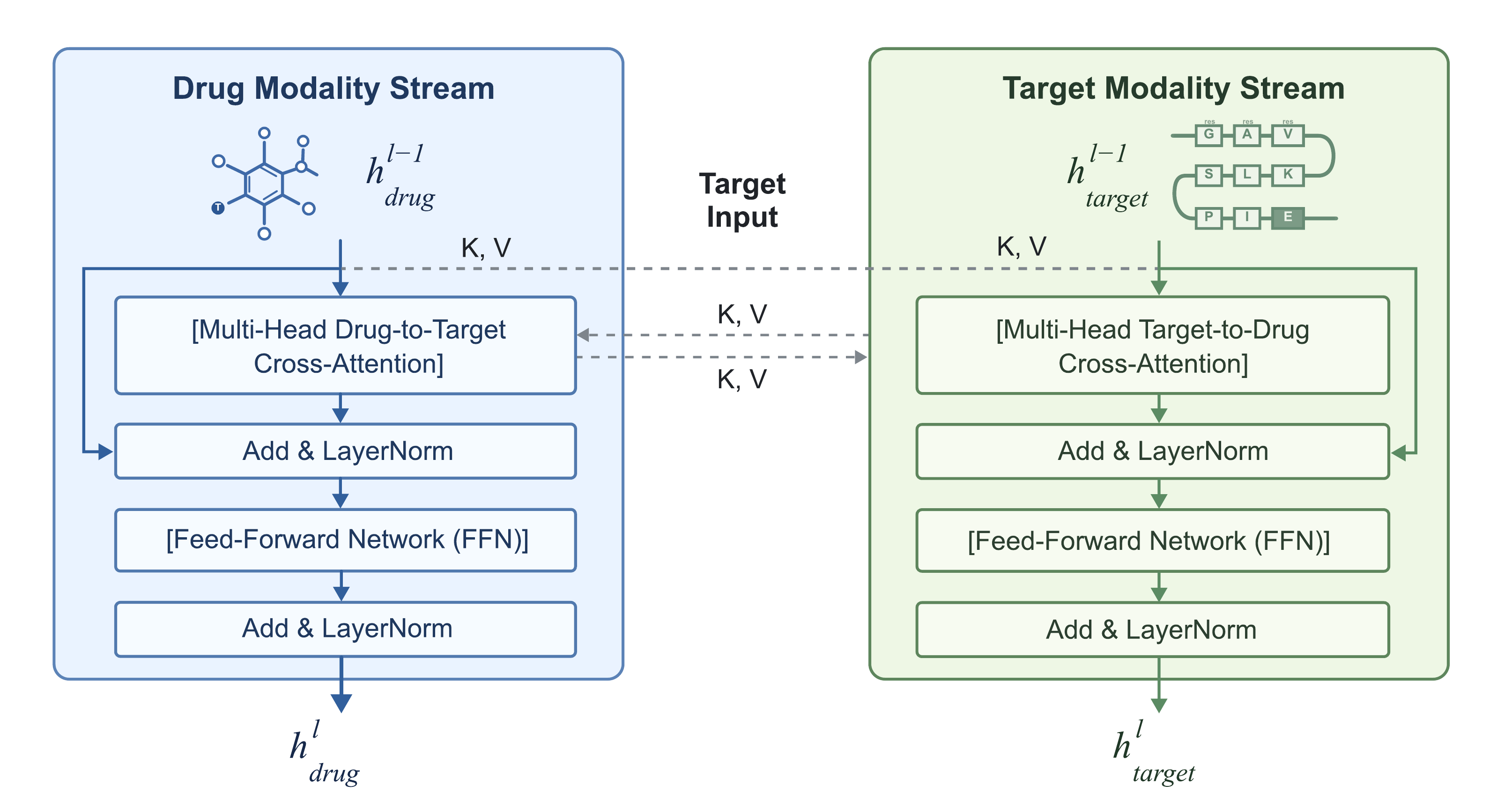}
\caption{Structure of the Cross-modal Joint Interaction Module.}
\label{fig:cjim}
\end{figure}

After projection and target fusion, the drug and target vectors have the same dimensionality but remain modality-specific. DPTM-DT uses a Cross-modal Joint Interaction Module (CJIM) to exchange information between them. The module follows the scaled dot-product attention principle introduced in Transformer models and adapts it to drug-target cross-modal matching \cite{attention,multimodaltransformer}. In one direction, the drug representation is used as the query and the target representation as key/value; in the opposite direction, the target representation queries the drug representation. For one attention head, the drug-to-target update is
\begin{equation}
A_{drug\leftarrow target}=
\mathrm{softmax}\left(\frac{Q_{drug}K_{target}^{T}}{\sqrt{d_h}}\right)V_{target}.
\end{equation}
The reverse path is defined symmetrically. Fig.~\ref{fig:cjim} shows the two-stream CJIM layer, where drug-to-target and target-to-drug cross-attention are followed by residual normalization and feed-forward updates. The reported configuration uses two interaction layers, eight attention heads, a model dimension of 256, and a feed-forward dimension of 1024.

\begin{figure}[t]
\centering
\includegraphics[width=0.96\columnwidth]{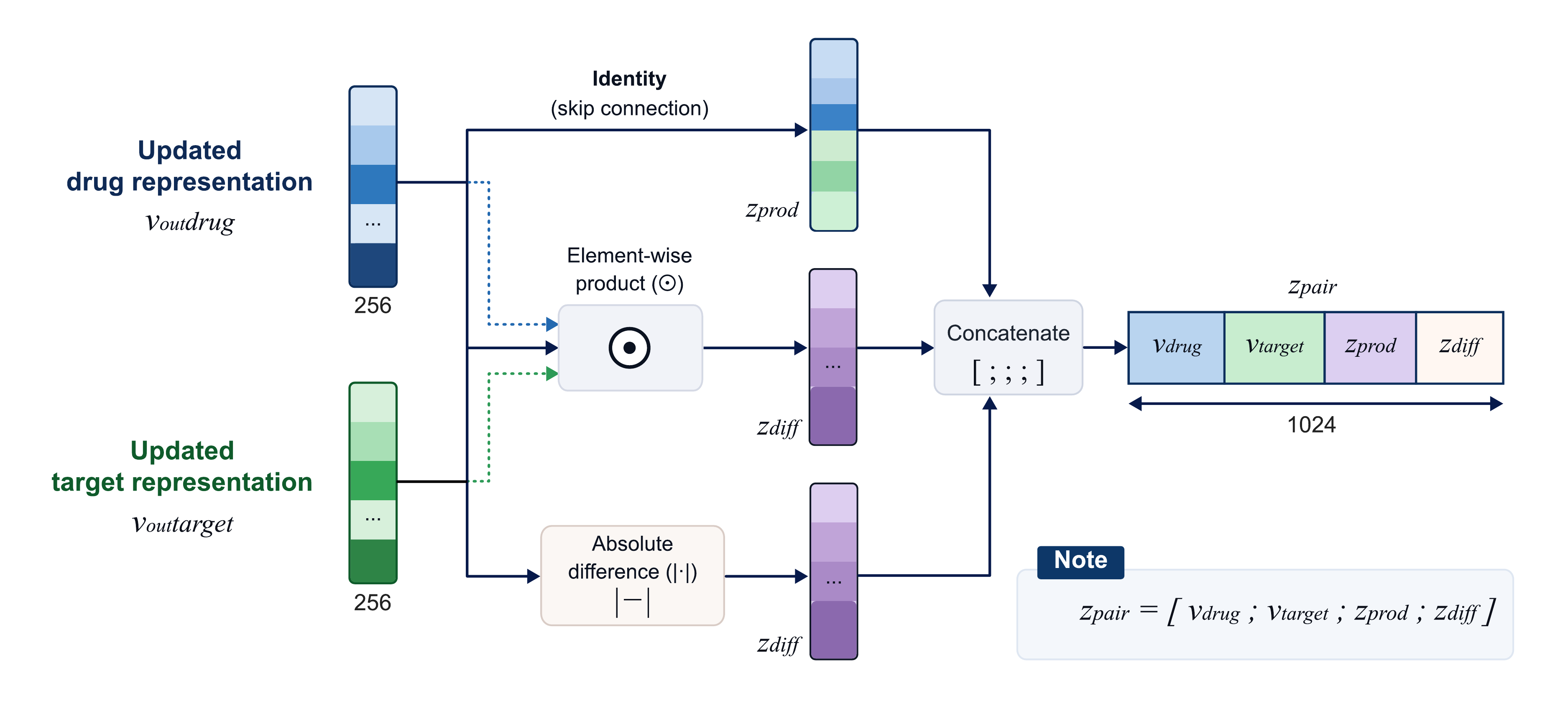}
\caption{Construction of the pair representation after cross-modal interaction. The updated drug and target vectors are concatenated with their element-wise product and absolute difference to form the final drug-target representation used by the prediction heads.}
\label{fig:pair_feature}
\end{figure}

The final pair vector concatenates four components:
\begin{equation}
z_i=[\tilde v_i^{drug};\tilde v_i^{target};
\tilde v_i^{drug}\odot\tilde v_i^{target};
|\tilde v_i^{drug}-\tilde v_i^{target}|].
\end{equation}
The product term describes local agreement between drug and target features, while the absolute difference term encodes feature mismatch. Fig.~\ref{fig:pair_feature} illustrates this construction, where the two 256-dimensional updated modality vectors are combined into a 1024-dimensional pair representation. This multiplicative component is consistent with compact bilinear-style interaction features used in multimodal representation learning \cite{hadamard}. A shared multilayer perceptron maps $z_i$ into a task-shared representation.

\subsection{Task Heads and Losses}

\begin{figure}[t]
\centering
\includegraphics[width=0.96\columnwidth]{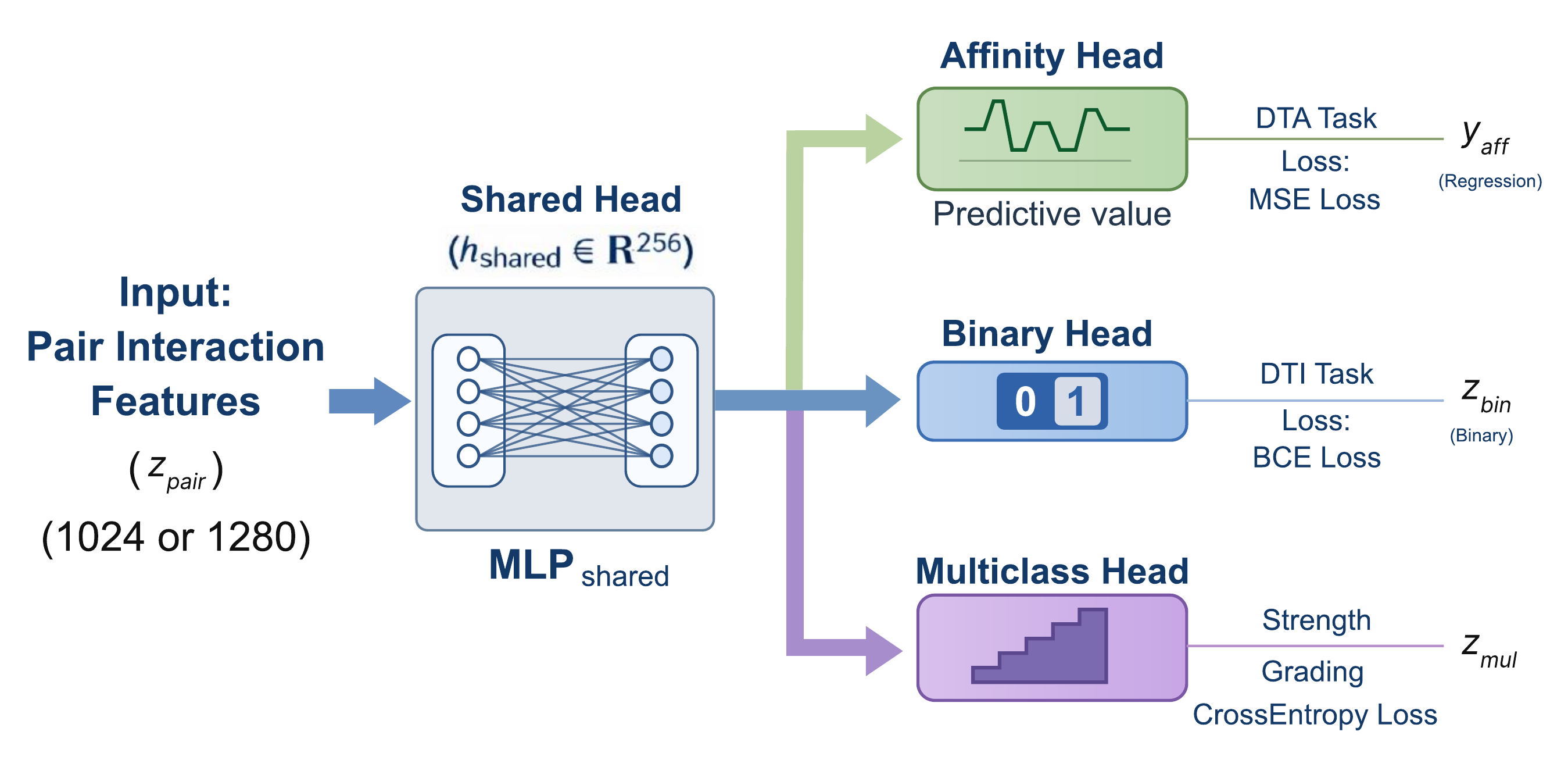}
\caption{Multitask prediction heads and training objectives.}
\label{fig:task_heads}
\end{figure}

The regression head predicts continuous affinity. The binary head predicts whether the affinity exceeds a dataset-specific threshold:
\begin{equation}
y_i^{bin}=\mathbb{I}(a_i\geq \tau),
\end{equation}
with $\tau=7.0$ for Davis pKd and $\tau=12.1$ for the KIBA score. The multiclass head predicts one of six affinity levels produced by Elbow-method-guided K-means discretization. For each training split, the continuous training affinities are clustered in the one-dimensional label space. The number of clusters is selected by the Elbow method and set to $C=6$ for both Davis and KIBA. The resulting cluster centers are then sorted by affinity strength, so the class index encodes an ordered low-to-high binding level:
\begin{equation}
y_i^{mul}=\mathrm{rank}\left(\arg\min_{c\in\{1,\ldots,C\}} |a_i-\mu_c|\right),
\end{equation}
where $\mu_c$ is the center of cluster $c$. This construction maps a continuous regression benchmark into an affinity-level recognition task; it does not replace the original regression objective. As illustrated in Fig.~\ref{fig:task_heads}, the shared pair representation is first mapped by a shared multilayer perceptron and then passed to task-specific heads. Training uses mean squared error for regression, binary cross-entropy with logits for binary classification, and cross-entropy for multiclass classification. AdamW with warmup and cosine learning-rate scheduling is used in training \cite{adamw}.

\section{Experiments and Results}

\subsection{Experimental Settings}

The experiments use Davis and KIBA, two common kinase-inhibitor benchmarks \cite{davisdata,kiba}. Davis contains 68 drugs, 442 targets, and 30,056 valid pairs. Its original $K_d$ values are converted to pKd, with a label range of 5.00--9.94, mean 5.45, and standard deviation 0.89. KIBA contains 2,111 drugs, 229 targets, and 118,254 valid pairs. Its score range is 0.00--17.20, with mean 11.72 and standard deviation 0.84. Binary labels use pKd $\geq 7.0$ for Davis and score $\geq 12.1$ for KIBA, producing positive ratios of 8.2\% and 21.0\%, respectively.

\begin{table}[t]
\caption{Dataset, Label, and Task Summary}
\label{tab:data}
\centering
\scriptsize
\begin{tabular}{lcc}
\toprule
Item & Davis & KIBA\\
\midrule
Drugs & 68 & 2,111\\
Targets & 442 & 229\\
Valid pairs & 30,056 & 118,254\\
Continuous label & pKd & KIBA score\\
Label range & 5.00--9.94 & 0.00--17.20\\
Mean / std. & 5.45 / 0.89 & 11.72 / 0.84\\
Positive threshold & pKd $\geq$ 7.0 & score $\geq$ 12.1\\
Positive ratio & 8.2\% & 21.0\%\\
Splits & \multicolumn{2}{c}{random 80/20 and standard}\\
Tasks & \multicolumn{2}{c}{regression, binary, six-class}\\
\bottomrule
\end{tabular}
\end{table}

Table~\ref{tab:tasks} summarizes how the three outputs are constructed and how they can be used in a screening workflow. The regression score supports fine-grained ranking, the binary output supports direct high-affinity filtering, and the six-class output provides an interpretable affinity-level label. The six classes are induced by the affinity distribution of each dataset and split rather than by manually fixed intervals, which is why both Macro-F1 and Weighted-F1 are reported for the multiclass task.

\begin{table}[t]
\caption{Task Outputs, Label Construction, and Screening Role}
\label{tab:tasks}
\centering
\scriptsize
\setlength{\tabcolsep}{4pt}
\begin{tabular}{p{0.16\columnwidth}p{0.36\columnwidth}p{0.36\columnwidth}}
\toprule
Output & Label construction & Main screening role\\
\midrule
Regression & Original continuous pKd or KIBA score & Rank candidate pairs and refine affinity estimates\\
Binary & pKd $\geq 7.0$ for Davis; KIBA score $\geq 12.1$ for KIBA & Filter likely high-affinity interactions\\
Six-class & Elbow-guided K-means on continuous affinities, with $C=6$ and centers sorted from low to high affinity & Provide coarse affinity-level triage and communicate binding strength\\
\bottomrule
\end{tabular}
\end{table}

Two split protocols are reported. The random 80/20 split supports model development and direct module comparison. The standard split follows the DeepDTA-style benchmark protocol and is used to reduce dependence on a single random partition. All comparisons are organized by the same dataset and split setting whenever the corresponding baseline results are available. The standard split is included specifically to make the comparison more referable to DeepDTA-style benchmark protocols, while avoiding a stronger claim that every external baseline uses identical internal feature preprocessing \cite{validationguidelines}. Regression is evaluated by MSE, RMSE, $R^2$, Pearson correlation, and concordance index (CI), where CI measures ranking consistency \cite{harrell}. Binary classification is evaluated by AUROC, AUPR, F1, and MCC, with AUPR and MCC being important under class imbalance \cite{matthews,davisgoadrich,saito2015,dtieval}. Multiclass prediction is evaluated by Accuracy, Macro-F1, Weighted-F1, and Top-2 Accuracy, which are commonly used to separate overall correctness from class-balanced behavior \cite{metricsstats}.

\subsection{Evaluation Details}

For regression, the affinity labels are standardized during training and transformed back to the original scale for evaluation. Feature and label normalization statistics are computed from the training partition only and then applied to validation and test partitions, which avoids leakage from the held-out data.

The model-selection metric is task-specific. The regression checkpoint is selected by validation MSE, the binary checkpoint by validation AUPR, and the multiclass checkpoint by validation Macro-F1. This choice follows the objective of each task: value fitting for regression, positive-pair retrieval under imbalance for binary classification, and class-balanced performance for the six-level task. The classification and multiclass results are reported with mean and standard deviation, reflecting repeated runs rather than a single random draw. Because paired per-seed predictions are not available for all baselines, formal statistical significance is not claimed beyond the reported repeated-run variability \cite{metricsstats}; small numerical differences are therefore interpreted conservatively.

The three tasks stress the representation in different ways. Regression measures continuous affinity estimation. Binary classification asks whether the learned representation still separates high-affinity pairs after thresholding. Six-class prediction tests whether the affinity scale can be divided into ordered binding levels. This gives a more complete view than a single regression score.

\subsection{Affinity Regression Results}

Affinity regression is the main continuous prediction task. Table~\ref{tab:regression} reports the regression comparison on Davis and KIBA. Under the random 80/20 split, DPTM-DT reaches MSE/CI values of 0.193/0.917 on Davis and 0.120/0.918 on KIBA. Under the standard split, it reports 0.195/0.913 on Davis and 0.124/0.906 on KIBA. In both split protocols and on both datasets, DPTM-DT has the lowest MSE among the compared methods.

\begin{table*}[t]
\caption{Affinity Regression Results on Davis and KIBA}
\label{tab:regression}
\centering
\scriptsize
\setlength{\tabcolsep}{4pt}
\begin{tabular}{llrrrrrrrr}
\toprule
Split & Model & \multicolumn{2}{c}{Davis} & & \multicolumn{2}{c}{KIBA} \\
\cmidrule{3-4}\cmidrule{6-7}
 & & CI$\uparrow$ & MSE$\downarrow$ & & CI$\uparrow$ & MSE$\downarrow$\\
\midrule
80/20 & DPTM-DT & 0.917 & 0.193 & & 0.918 & 0.120\\
80/20 & TEFDTA & 0.895 & 0.194 & & 0.865 & 0.179\\
80/20 & MFR-DTA & 0.910 & 0.216 & & 0.903 & 0.131\\
80/20 & GLCN-DTA & 0.908 & 0.210 & & 0.878 & 0.122\\
80/20 & DeepDTA & 0.883 & 0.256 & & 0.868 & 0.189\\
80/20 & WideDTA & 0.891 & 0.257 & & 0.880 & 0.174\\
\midrule
Standard & DPTM-DT & 0.913 & 0.195 & & 0.906 & 0.124\\
Standard & TEFDTA & 0.891 & 0.198 & & 0.858 & 0.183\\
Standard & MFR-DTA & 0.905 & 0.226 & & 0.896 & 0.134\\
Standard & GLCN-DTA & 0.905 & 0.217 & & 0.874 & 0.128\\
Standard & DeepDTA & 0.883 & 0.260 & & 0.855 & 0.192\\
Standard & WideDTA & 0.885 & 0.264 & & 0.868 & 0.176\\
\bottomrule
\end{tabular}
\end{table*}

The regression pattern is consistent. On Davis, DPTM-DT slightly improves MSE over TEFDTA while obtaining a clearer CI advantage, which indicates better ranking of candidate pairs. On KIBA, DPTM-DT is close to GLCN-DTA in MSE under the 80/20 split but has a much higher CI. Because DTA screening often requires both accurate values and reliable ordering, this paired improvement is important. The standard-split results remain close to the random-split results, suggesting that the model is not overly dependent on one favorable random split.

The regression table also shows why reporting only one metric is incomplete. MSE reflects value-level error, while CI reflects whether the model preserves the relative order of affinities. DPTM-DT is competitive in both senses. For example, on KIBA under the 80/20 split, the MSE difference between DPTM-DT and GLCN-DTA is small, but the CI difference is larger. This indicates that the proposed representation is useful for both fitting affinity scores and prioritizing candidate pairs in virtual screening.

\subsection{Binary Classification Results}

The binary task converts continuous affinity labels into interaction labels using the thresholds in Table~\ref{tab:data}. This setting is more imbalanced than the regression task, especially for Davis, where positive samples are only 8.2\%. Table~\ref{tab:binary} organizes the binary results by split and dataset.

\begin{table*}[t]
\caption{Binary Classification Results}
\label{tab:binary}
\centering
\scriptsize
\setlength{\tabcolsep}{3pt}
\begin{tabular}{lllcccc}
\toprule
Split & Dataset & Model & AUROC$\uparrow$ & AUPR$\uparrow$ & F1$\uparrow$ & MCC$\uparrow$\\
\midrule
80/20 & Davis & DPTM-DT & 0.948$\pm$0.006 & 0.727$\pm$0.024 & 0.681$\pm$0.028 & 0.654$\pm$0.022\\
80/20 & Davis & TEFDTA & 0.931$\pm$0.004 & 0.698$\pm$0.011 & 0.655$\pm$0.011 & 0.622$\pm$0.027\\
80/20 & Davis & MFR-DTA & 0.941$\pm$0.008 & 0.718$\pm$0.020 & 0.670$\pm$0.011 & 0.640$\pm$0.029\\
80/20 & Davis & GLCN-DTA & 0.928$\pm$0.010 & 0.692$\pm$0.012 & 0.648$\pm$0.015 & 0.615$\pm$0.014\\
80/20 & Davis & DeepDTA & 0.915$\pm$0.006 & 0.680$\pm$0.017 & 0.632$\pm$0.021 & 0.598$\pm$0.016\\
80/20 & Davis & WideDTA & 0.921$\pm$0.009 & 0.688$\pm$0.010 & 0.640$\pm$0.017 & 0.608$\pm$0.017\\
\midrule
80/20 & KIBA & DPTM-DT & 0.956$\pm$0.007 & 0.798$\pm$0.021 & 0.714$\pm$0.015 & 0.689$\pm$0.020\\
80/20 & KIBA & TEFDTA & 0.942$\pm$0.008 & 0.762$\pm$0.009 & 0.685$\pm$0.025 & 0.658$\pm$0.013\\
80/20 & KIBA & MFR-DTA & 0.950$\pm$0.004 & 0.782$\pm$0.024 & 0.702$\pm$0.034 & 0.675$\pm$0.026\\
80/20 & KIBA & GLCN-DTA & 0.940$\pm$0.006 & 0.758$\pm$0.010 & 0.680$\pm$0.027 & 0.652$\pm$0.019\\
80/20 & KIBA & DeepDTA & 0.930$\pm$0.004 & 0.745$\pm$0.016 & 0.665$\pm$0.011 & 0.635$\pm$0.028\\
80/20 & KIBA & WideDTA & 0.936$\pm$0.005 & 0.752$\pm$0.019 & 0.672$\pm$0.018 & 0.645$\pm$0.020\\
\midrule
Std. & Davis & DPTM-DT & 0.895$\pm$0.008 & 0.709$\pm$0.011 & 0.659$\pm$0.034 & 0.630$\pm$0.026\\
Std. & Davis & MFR-DTA & 0.885$\pm$0.011 & 0.705$\pm$0.023 & 0.652$\pm$0.025 & 0.622$\pm$0.028\\
Std. & Davis & TEFDTA & 0.874$\pm$0.004 & 0.674$\pm$0.011 & 0.634$\pm$0.011 & 0.598$\pm$0.017\\
Std. & KIBA & DPTM-DT & 0.908$\pm$0.006 & 0.776$\pm$0.013 & 0.696$\pm$0.031 & 0.664$\pm$0.017\\
Std. & KIBA & MFR-DTA & 0.894$\pm$0.006 & 0.768$\pm$0.017 & 0.686$\pm$0.014 & 0.658$\pm$0.026\\
Std. & KIBA & TEFDTA & 0.882$\pm$0.004 & 0.741$\pm$0.025 & 0.668$\pm$0.029 & 0.638$\pm$0.014\\
\bottomrule
\end{tabular}
\end{table*}

Under the 80/20 split, DPTM-DT obtains the highest AUROC, AUPR, F1, and MCC on both Davis and KIBA. On Davis, AUPR increases from 0.718 for the strongest listed baseline to 0.727, and MCC increases from 0.640 to 0.654. On KIBA, AUPR increases from 0.782 to 0.798. Scores decrease under the standard split, as expected, but DPTM-DT remains ahead of the strongest reported baselines. The binary task is useful here because it tests high-affinity recognition directly, rather than only the fit to continuous affinity values.

The binary task is sensitive to class imbalance, especially on Davis, where the negative class dominates. For this reason, AUPR is more informative than accuracy for positive-pair retrieval. MCC complements it by using all four entries of the confusion matrix. DPTM-DT improves both metrics, which indicates better positive retrieval without losing the balance between positive and negative predictions.

\subsection{Multiclass Classification Results}

The multiclass task maps continuous affinity labels into six ordered categories. It complements standard affinity regression by evaluating whether the model can distinguish affinity levels. The classes are generated from the continuous label distribution by the K-means procedure described in the Method section, with class indices ordered according to cluster centers. This means that class frequencies are dataset-dependent rather than manually balanced. Table~\ref{tab:multiclass} separates the random and standard split results and reports the main classification metrics.

\begin{table*}[t]
\caption{Six-Class Affinity-Level Classification Results}
\label{tab:multiclass}
\centering
\scriptsize
\setlength{\tabcolsep}{3pt}
\begin{tabular}{lllcccc}
\toprule
Split & Dataset & Model & Acc.$\uparrow$ & Macro-F1$\uparrow$ & W-F1$\uparrow$ & Top-2$\uparrow$\\
\midrule
80/20 & Davis & DPTM-DT & 0.826$\pm$0.006 & 0.800$\pm$0.020 & 0.819$\pm$0.008 & 0.932$\pm$0.006\\
80/20 & Davis & TEFDTA & 0.812$\pm$0.006 & 0.759$\pm$0.017 & 0.799$\pm$0.009 & 0.911$\pm$0.003\\
80/20 & Davis & MFR-DTA & 0.814$\pm$0.008 & 0.760$\pm$0.015 & 0.804$\pm$0.007 & 0.917$\pm$0.004\\
80/20 & Davis & GLCN-DTA & 0.797$\pm$0.007 & 0.752$\pm$0.022 & 0.783$\pm$0.007 & 0.894$\pm$0.007\\
80/20 & Davis & DeepDTA & 0.772$\pm$0.004 & 0.733$\pm$0.025 & 0.765$\pm$0.010 & 0.892$\pm$0.007\\
80/20 & Davis & WideDTA & 0.781$\pm$0.006 & 0.722$\pm$0.019 & 0.768$\pm$0.008 & 0.878$\pm$0.007\\
\midrule
80/20 & KIBA & DPTM-DT & 0.847$\pm$0.006 & 0.815$\pm$0.024 & 0.840$\pm$0.010 & 0.962$\pm$0.004\\
80/20 & KIBA & TEFDTA & 0.831$\pm$0.009 & 0.792$\pm$0.018 & 0.819$\pm$0.011 & 0.953$\pm$0.006\\
80/20 & KIBA & MFR-DTA & 0.836$\pm$0.008 & 0.783$\pm$0.020 & 0.829$\pm$0.010 & 0.960$\pm$0.007\\
80/20 & KIBA & GLCN-DTA & 0.816$\pm$0.008 & 0.767$\pm$0.020 & 0.805$\pm$0.008 & 0.924$\pm$0.006\\
80/20 & KIBA & DeepDTA & 0.794$\pm$0.005 & 0.750$\pm$0.028 & 0.787$\pm$0.009 & 0.914$\pm$0.005\\
80/20 & KIBA & WideDTA & 0.808$\pm$0.007 & 0.760$\pm$0.023 & 0.796$\pm$0.010 & 0.911$\pm$0.004\\
\midrule
Std. & Davis & DPTM-DT & 0.816$\pm$0.006 & 0.782$\pm$0.022 & 0.808$\pm$0.010 & 0.923$\pm$0.006\\
Std. & Davis & TEFDTA & 0.788$\pm$0.006 & 0.743$\pm$0.023 & 0.776$\pm$0.011 & 0.903$\pm$0.004\\
Std. & Davis & MFR-DTA & 0.802$\pm$0.009 & 0.754$\pm$0.019 & 0.797$\pm$0.011 & 0.918$\pm$0.004\\
Std. & Davis & GLCN-DTA & 0.778$\pm$0.009 & 0.739$\pm$0.022 & 0.768$\pm$0.007 & 0.893$\pm$0.004\\
Std. & Davis & DeepDTA & 0.758$\pm$0.006 & 0.718$\pm$0.024 & 0.743$\pm$0.011 & 0.863$\pm$0.005\\
Std. & Davis & WideDTA & 0.772$\pm$0.009 & 0.732$\pm$0.015 & 0.765$\pm$0.006 & 0.886$\pm$0.006\\
Std. & KIBA & DPTM-DT & 0.833$\pm$0.008 & 0.799$\pm$0.026 & 0.827$\pm$0.011 & 0.942$\pm$0.003\\
Std. & KIBA & TEFDTA & 0.807$\pm$0.007 & 0.755$\pm$0.025 & 0.800$\pm$0.008 & 0.902$\pm$0.006\\
Std. & KIBA & MFR-DTA & 0.813$\pm$0.005 & 0.758$\pm$0.022 & 0.804$\pm$0.007 & 0.935$\pm$0.005\\
Std. & KIBA & GLCN-DTA & 0.795$\pm$0.007 & 0.748$\pm$0.019 & 0.787$\pm$0.010 & 0.897$\pm$0.003\\
Std. & KIBA & DeepDTA & 0.775$\pm$0.009 & 0.731$\pm$0.024 & 0.763$\pm$0.005 & 0.873$\pm$0.003\\
Std. & KIBA & WideDTA & 0.782$\pm$0.005 & 0.740$\pm$0.015 & 0.774$\pm$0.008 & 0.885$\pm$0.003\\
\bottomrule
\end{tabular}
\end{table*}

The multiclass results show the same ordering trend as the regression and binary experiments. DPTM-DT obtains the best Macro-F1 and Accuracy in every reported setting, reaching Macro-F1 values of 0.800/0.815 under the 80/20 split and 0.782/0.799 under the standard split on Davis/KIBA. KIBA is consistently higher than Davis, which matches the larger number of valid pairs and higher positive ratio in KIBA. The reported Top-2 Accuracy is much higher than exact Accuracy, indicating that many errors remain close to the correct affinity level. This pattern suggests that the six-class task mainly evaluates affinity-level ordering and neighborhood confusion.

The six-class task gives another view of the continuous labels. Macro-F1 measures performance across affinity levels, whereas Weighted-F1 reflects the class distribution induced by discretization. Top-2 Accuracy is useful because adjacent affinity bins can be hard to separate sharply. Across both datasets, DPTM-DT keeps the best exact and Top-2 performance in the reported comparisons, supporting the use of one shared pair representation for both numerical prediction and coarser affinity-level decisions.

\subsection{Ablation Study}

Table~\ref{tab:ablation} summarizes representative ablations on Davis under the 80/20 split across the three tasks. The regression ablation reports all five regression metrics for the full model and the main variants. Binary and multiclass columns keep the metrics most directly tied to the classification objectives.

\begin{table*}[t]
\caption{Representative Davis 80/20 Ablation Results}
\label{tab:ablation}
\centering
\scriptsize
\setlength{\tabcolsep}{3pt}
\begin{tabular}{lccccc|cc|cc}
\toprule
Variant & MSE$\downarrow$ & RMSE$\downarrow$ & $R^2\uparrow$ & Pearson$\uparrow$ & CI$\uparrow$ & AUPR$\uparrow$ & MCC$\uparrow$ & Macro-F1$\uparrow$ & Top-2$\uparrow$\\
\midrule
DPTM-DT & 0.193 & 0.440 & 0.754 & 0.869 & 0.917 & 0.727 & 0.654 & 0.800 & 0.932\\
ESM-only & 0.199 & 0.446 & 0.753 & 0.866 & 0.906 & 0.680 & 0.610 & 0.584 & 0.818\\
CTD-only & 0.216 & 0.465 & 0.726 & 0.853 & 0.886 & 0.675 & 0.605 & 0.538 & 0.772\\
Concat & 0.198 & 0.445 & 0.752 & 0.867 & 0.897 & 0.722 & 0.645 & 0.777 & 0.928\\
No cross-attn & 0.200 & 0.447 & 0.750 & 0.865 & 0.892 & 0.715 & 0.638 & 0.773 & 0.925\\
\bottomrule
\end{tabular}
\end{table*}

The ablation results separate the roles of the main modules. ESM-only is consistently stronger than CTD-only, so the pretrained protein embedding is the main target-side signal. The full model still improves over both single-source variants, showing that CTD statistics add useful physicochemical information. Gated fusion is also consistently better than direct concatenation, which suggests that the target branch benefits from adaptive feature weighting. Finally, removing cross-attention weakens regression, binary classification, and multiclass classification, indicating that drug-target information exchange remains useful after unimodal representation learning.

These ablations also help interpret the relative roles of DCRM and CJIM. DCRM improves the completeness of the target representation by combining deep sequence semantics with explicit physicochemical statistics. CJIM then uses these representations to form a pair-aware interaction signal. The No-cross-attention variant keeps the same input features but weakens the interaction mechanism, and its drop across all three task groups indicates that the model benefits from explicit cross-modal exchange rather than only from stronger unimodal encoders. Although the margin size varies by metric, the direction is stable across regression, binary classification, and multiclass classification.

\subsection{Cross-Task Discussion}

Across the experiments, DPTM-DT shows consistent cross-task behavior. The same architectural components improve continuous affinity prediction, binary interaction recognition, and affinity-level classification. This matters because the three tasks come from the same affinity labels but evaluate different aspects of the relation: numerical error and ranking, high-affinity detection under imbalance, and ordinal binding-strength levels.

The results are also stable across split protocols. Scores under the standard split are lower than those under the random 80/20 split, but DPTM-DT keeps its relative advantage in every task. KIBA generally gives stronger absolute scores than Davis, which is consistent with its larger sample size and less extreme positive-class imbalance.

The experimental trends match the architecture in Fig.~\ref{fig:framework} and the pair-feature construction in Fig.~\ref{fig:pair_feature}. Improvements appear across tasks rather than in a single output head. This supports the main design claim: pretrained molecular and protein representations are more effective for drug-target prediction when they are connected through adaptive target fusion, explicit cross-modal exchange, and task-shared pair modeling.

\subsection{Practical Interpretation}

The three outputs can be interpreted as a compact decision pipeline. The regression output provides a continuous score for ranking candidate pairs. The binary output gives a direct high-affinity decision using dataset-specific thresholds. The six-class output gives a coarser affinity-level label that can be easier to communicate in an interface or downstream filtering workflow. In an analysis system, a user can enter a drug SMILES string and a protein FASTA sequence and receive an interaction judgment, an affinity estimate, and an affinity-level prediction. This use case clarifies why the three-task design is useful from an application perspective.

Regression remains the primary DTA evidence because Davis and KIBA are continuous-label benchmarks. The binary and six-class tasks are derived from the same labels and test whether the representation remains useful after discretization. They enrich the evaluation, but they should be read as additional task views rather than independent datasets.

The evaluation is centered on two widely used kinase-inhibitor benchmarks and two split settings, which makes the results directly comparable with common DTA protocols. This benchmark-centered scope follows the validation practice of many drug-target prediction studies while keeping the claims tied to measurable prediction performance and module-level evidence \cite{validationguidelines,dtieval}. Within that scope, the experiments show that the same pair representation can support value prediction, high-affinity retrieval, and affinity-level discrimination.

\section{Conclusion}

This paper presents DPTM-DT, a dual-pretrained cross-modal multitask framework for drug-target prediction. The model combines GROVER drug embeddings, ESM protein embeddings, CTD target descriptors, gated target fusion, bidirectional cross-modal attention, and task-specific prediction heads over a shared pair representation. Experiments cover Davis and KIBA, random 80/20 and standard splits, affinity regression, binary classification, six-class affinity-level classification, and Davis 80/20 ablations. The regression head ranks candidate pairs by continuous affinity, the binary head filters high-affinity interactions, and the six-class head gives a coarse binding-strength level for downstream triage. Results favor DPTM-DT across tasks and splits, and the ablations support the roles of dual representation, adaptive fusion, and cross-modal interaction. Overall, DPTM-DT offers a practical multitask representation-learning framework for benchmark DTA prediction and can be extended to broader target families and model-behavior analysis.


\end{document}